\documentclass[cameraready]{Interspeech}
\title{Measuring the Redundancy of Decoder Layers in SpeechLLMs\thanks{Adel Moumen is funded by the Cambridge Trust.}}

\author[]{Adel}{Moumen}
\author[]{Guangzhi}{Sun}
\author[]{Philip C}{Woodland}
\address{
    Department of Engineering, University of Cambridge, UK
}

\email{\{am3303,gs534,pcw117\}@cam.ac.uk}


\usepackage{comment}

\begin{document}
\maketitle

\begin{abstract}
  Speech Large Language Models route speech encoder representations into an LLM decoder that typically accounts for over 90\% of total parameters. We study how much of this decoder capacity is actually needed for speech tasks. Across two LLM families and three scales (1--8B), we show that decoder redundancy is largely inherited from the pretrained LLM: text and speech inputs yield similar redundant blocks. We then measure excess capacity by pruning decoder layers and analysing post-pruning healing to increase robustness. Our findings show that 7--8B models retain good ASR performance with only $\sim$60\% of decoder layers, and the same trend extends to smaller scales with reduced pruning tolerance. We then generalise to speech translation, and show that the same blocks of layers are redundant across speech encoders, tasks and languages, indicating that a more global redundancy structure exists, enabling a single pruned SpeechLLM backbone to support multiple tasks at lower computational cost.
\end{abstract}

\section{Introduction}

Speech Large Language Models (SpeechLLMs) \cite{ma2024embarrassingly, 10445874, saon2025granite,arora2025landscape} combine a speech encoder, a projector, and a pretrained Large Language Model (LLM) decoder to perform speech tasks such as Automatic Speech Recognition (ASR) \cite{fathullah2024prompting,mittal2024salsa, hori2025delayed, hsu2024let} and Automatic Speech Translation (AST) \cite{pan2023cosmic,wu2023decoder}. This approach has gained popularity and now achieves state-of-the-art results \cite{srivastav2025openasrleaderboardreproducible}.

However, the decoder typically dominates the parameter budget, often exceeding 90\% of the total \cite{ma2024embarrassingly, saon2025granite}, yet speech tasks have traditionally been addressed with substantially smaller models \cite{radford2023robust, moumen2023stabilising}. This raises a natural question: \emph{How much of this decoder capacity is actually needed for speech tasks?}

We approach this question through the lens of layer redundancy: if layers can be removed with small downstream performance degradation, then the decoder has more capacity than the task requires. Quantifying this redundancy is valuable both practically, as identifying redundancy enables smaller and faster task-specific decoders, and scientifically, as it reveals whether SpeechLLMs actually exploit their full capacity.

Prior work has shown that LLMs contain significant redundancy \cite{gromov2025unreasonableineffectivenessdeeperlayers, belrose2023eliciting, sajjad2023effect, yang2024laco, men2025shortgpt}. In the speech domain, speech encoders \cite{baevski2020wav2vec,chen2022wavlm} also exhibit substantial redundancy \cite{liu2021tera, peng2023structured, zhang2021usefulness}. Knowledge distillation \cite{gou2021knowledge, hinton2015distilling} has been used to compress ASR systems like Whisper \cite{radford2023robust}, removing 51\% of its layers \cite{gandhi2023distilwhisperrobustknowledgedistillation, sy2025baldwhisperfasterwhisperhead}, suggesting that even task-specific systems carry excess capacity. However, to our knowledge, redundancy in SpeechLLM decoders has not been systematically characterised, which leaves the question of excess capacity open.

We investigate decoder redundancy across two LLM families and three scales, focusing on ASR and then evaluating transfer to speech translation.
Our contributions are as follows:
\begin{itemize}
  \item We show that the decoder redundancy in SpeechLLMs is inherited from the pretrained LLM: under both text and speech, the decoder layers exhibit similar inter-layer patterns, and fine-tuning the decoder further amplifies this structure.
  \item We characterise how decoder redundancy scales with model size on ASR, showing that larger models are more prunable, with up to 43.8\% of decoder layers removable while retaining good ASR performance.
  \item We analyse the healing mechanism behind pruning and find that jointly adapting the decoder and projector is critical for pruning robustness.
  \item We transfer these findings to AST and show that ASR- and AST-optimal pruning layers closely coincide despite different tasks, source languages, and speech encoders, suggesting a broader redundancy phenomenon.
\end{itemize}
Code and checkpoints will be released through the SpeechBrain toolkit \cite{ravanelli2024open, ravanelli2021speechbrain}.
% The rest of the paper is organised as follows. Section~\ref{sec:speechllms} describes the SpeechLLM framework and Section~\ref{sec:redundancy} formalises our redundancy measure. Section~\ref{sec:experiments} details the experimental setup and Section~\ref{sec:results} presents the results.  
\section{Speech Large Language Models}
\label{sec:speechllms}
We focus on the SLAM recipe \cite{ma2024embarrassingly}, a popular SpeechLLM framework that composes a speech encoder, a lightweight projector, and a frozen LLM decoder. We describe each component and how they are combined.

\subsection{SLAM Framework}

\noindent \textbf{Speech encoder.} A speech encoder $f:\mathbb{R}^{T}\rightarrow\mathbb{R}^{T'\times d_e}$ of dimension $d_e$ maps a waveform $\mathbf{x}\in\mathbb{R}^{T}$ to a sequence of continuous feature vectors $\mathbf{S}=f(\mathbf{x})=(\mathbf{s}_{1},\dots,\mathbf{s}_{T'})$ with a much reduced sequence length $T'\ll T$. Such encoders are usually pretrained with self-supervised. Because speech sequences are typically long, $k$ consecutive frames are concatenated along the feature dimension, yielding $\mathbf{Z}^{S}\in\mathbb{R}^{N\times kd_e}$ with $N=T'/k$.

\noindent \textbf{Projector.} Speech encoders and LLMs are usually designed and trained independently. As a result, their internal representation spaces rarely align naturally. To bridge this gap, a projector is introduced to convert the downsampled speech representations into the LLM embedding space. This is a single-hidden-layer MLP with a non-linear GELU activation in the middle:
\begin{equation}
\mathbf{E}^{S}=\mathrm{Linear}\big(\mathrm{GELU}(\mathrm{Linear}(\mathbf{Z}^{S}))\big).
\label{eq:slam_projector}
\end{equation}
In the SLAM recipe, the projector is the only trained component, learning to align the speech representations with the decoder's internal representations.

\noindent \textbf{LLM decoder.} The input sequence $\mathbf{E}=(\mathbf{E}^{S},\mathbf{E}^{P},\mathbf{E}^{G})$ concatenates speech, prompt, and target embeddings. Then, a decoder-only LLM consisting of $m$ causal transformer layers maps $\mathbf{E}$ to contextual hidden states $(\mathbf{h}_{1},\dots,\mathbf{h}_{L})$ where $L$ corresponds to the input sequence length. Training uses teacher forcing, minimising the negative log-likelihood over the target tokens. In practice, $m$ can be large (e.g.\ $32$ for Llama3.1-8B), making the decoder the dominant parameter cost.

\noindent \textbf{Fine-tuning.} In the standard SLAM recipe, the decoder remains frozen and only the projector is trained. A common variant applies Low-Rank Adapters (LoRA) \cite{hu2022lora} to the decoder's attention projections, enabling lightweight adaptation while keeping most parameters fixed.

\section{Measuring Decoder Redundancy}
\label{sec:redundancy}
LLM decoders are typically deep, and it is unclear whether all layers are needed for speech tasks. We formalise a measure of redundancy based on angular distance and describe the pruning algorithm used to quantify the redundancy following \cite{gromov2025unreasonableineffectivenessdeeperlayers}. %Unlike knowledge distillation \cite{gandhi2023distilwhisperrobustknowledgedistillation}, which requires training a separate student model, the angular distance pruning method is computed from a single forward pass on the model, making it more practical for comparison across models.

\subsection{Redundancy proxy}
The decoder redundancy is measured as the largest fraction of contiguous layers that can be removed with only a small degradation on the downstream metric. To identify removable blocks efficiently, we follow \cite{gromov2025unreasonableineffectivenessdeeperlayers} and use the angular distance between hidden representations as a proxy. For a candidate block of $n$ layers starting at layer $\ell$, we compute the angular distance $d(\mathbf{h}_{\ell}, \mathbf{h}_{\ell+n})$ between the hidden states $\mathbf{h}_{\ell}$ and $\mathbf{h}_{\ell+n}$, averaged over task-representative inputs (e.g. validation set). Here, $\mathbf{h}_{\ell}$ is the hidden state extracted for the last token in the sequence.

\subsection{Pruning algorithm}
For each block size $n$, the start layer is selected that minimises angular distance:
\begin{equation}
  \ell^*(n) = \arg\min_{\ell \in \{1,\dots,m-n\}} d(\mathbf{h}_{\ell},\mathbf{h}_{\ell+n}).
\end{equation}
% todo: fix capital L
Given block size $n$, we remove layers $\ell^*{+}1,\dots,\ell^*{+}n{-}1$ and connect the output of layer $\ell^*$ directly to the $(\ell^*{+}n)$-th block. The sequence $\{\ell^*(n)\}_{n=2}^{m-1}$ defines the \emph{optimal pruning path} for a given model and task: each point specifies which block to remove for a given pruning depth. % The block size $n$ tells us the pair of layers that are the most compatible. 

\subsection{Post-pruning healing}
Low angular distance indicates that two representations are geometrically close, but does not guarantee compatibility: the receiving layer $\ell$ expects input from its predecessor $\ell-1$, not from a distant earlier layer. Without compensation, even removing blocks with small angular distance can cause sharp degradation. To bridge this gap, we \emph{heal} the pruned interface by attaching LoRA adapters to the $(\ell^*{+}n)$-th block MLP (a small fraction of the total layer parameters), allowing it to learn the residual correction needed after the removed block. Additionally, we optionally unfreeze the projector: removing a block shifts the input distribution of the receiving layer $(\ell^*{+}n)$, so the projector's alignment becomes stale and is re-adapted to the new decoder dynamics.

\section{Experimental Setup}
\label{sec:experiments}
Redundancy is first studied on ASR using the SLAM framework, adding one LoRA-adapted configuration (Qwen2.5-7B) to test whether adaptation changes the picture. Three healing strategies are compared: decoder-only, projector-only, and both jointly. To assess generalisation beyond ASR, we run a task generalisation study on AST (En$\rightarrow$De and Fr$\rightarrow$En).

\noindent \textbf{Models.} WavLM Large \cite{chen2022wavlm} is used as the speech encoder, which has been shown to work well in SLAM-style systems \cite{ma2024embarrassingly, kumar2025performance}, with a downsampling factor of $k{=}5$ ($10$~Hz frame rate). For the Fr$\rightarrow$En AST experiments the Whisper Large v3 encoder is used as it provides multilingual representations. To study redundancy across architectures and scales, we consider two LLM families, Qwen2.5 \cite{yang2025qwen3} and Llama (3.1/3.2) \cite{grattafiori2024llama}, across three size regimes (1--1.5B, 3--4B, 7--8B), yielding six backbones with different depth/width profiles.

\noindent \textbf{Training.}
During training, only the projector is trained. We use 2$\times$ H100 (80~GiB) GPUs with a per-GPU batch size of $8$, Adam optimiser \cite{kingma2014adam} with $(\beta_1,\beta_2){=}(0.9,0.98)$, gradient clipping of $1.0$, and a peak learning rate of $5{\times}10^{-4}$ with linear warmdown ($60\%$). Models are trained for $50{,}000$ iterations ($\sim3$ epochs); Llama~3.1-8B uses $75{,}000$ with $80\%$ warmdown. The LoRA-adapted Qwen2.5-7B adds rank-$64$ adapters ($\alpha{=}64$, dropout $0.05$) to the decoder's $\mathbf{q}$, $\mathbf{k}$, $\mathbf{v}$, $\mathbf{o}$ projections, and all other hyperparameters are shared.

\noindent \textbf{Post-pruning healing.}
After pruning, we optionally heal the receiving decoder MLP with LoRA (rank $64$, dropout $0.05$) and/or unfreeze the projector for $5{,}000$ iterations on the same training data using the same hyperparameters as before.

\subsection{Datasets and evaluation}
\noindent \textbf{LibriSpeech} \cite{panayotov2015librispeech} consists of read English speech derived from audiobooks sampled at $16$~kHz. We train ASR models on the $960$~h partition and evaluate in-domain on test-clean/test-other.

\noindent \textbf{Loquacious} \cite{parcollet2025loquacious} aggregates diverse English speech sources \cite{ardila2020common,wang2021voxpopuli,li2023yodas} making it a challenging evaluation dataset. We evaluate on its dev partition for out-of-domain assessment, since its test set includes LibriSpeech test-clean and test-other.

\noindent \textbf{CoVoST2} \cite{wang2021covost} is a multilingual speech-to-text translation corpus. We focus on two language pairs: En$\rightarrow$De and Fr$\rightarrow$En, with training set durations of $428$ and $264$ hours respectively.

\noindent \textbf{Decoding and metrics.} Greedy decoding is used at inference time for all models. ASR is evaluated with WER, with the Whisper English text normaliser applied for consistency across datasets. For AST, BLEU scores are reported using SacreBLEU \cite{post-2018-call}. Pruning tolerance is summarised by relative degradation: given a baseline score $s_0$, the relative change is defined as $\Delta = (s - s_0) / s_0$. For ASR, the largest fraction of removable layers is reported such that $\Delta_{\mathrm{WER}} \leq 0.25$ across all evaluation sets simultaneously. The $0.25$ threshold matches the average relative WER gap between adjacent size tiers, so pruning does not degrade a model below the next smaller tier. For AST, a stricter threshold of $\Delta_{\mathrm{BLEU}} \leq 0.10$ is adopted, corresponding to at most a 10\% relative decrease in BLEU. This stricter threshold is chosen to ensure the absolute BLEU score remains close to the unpruned performance.
\section{Experimental Results}

% This threshold is based on Section~\ref{sec:results} results: within each model family, the average cumulative relative WER gap between adjacent size tiers (7--8B vs.\ 3--4B, and 3--4B vs.\ 1--1.5B) is approximately $0.25$. At this pruning level, WER is not degraded beyond the unpruned performance of the next smaller tier. 
\label{sec:results}

\begin{figure*}[!ht]
  \centering
  \includegraphics[width=\textwidth]{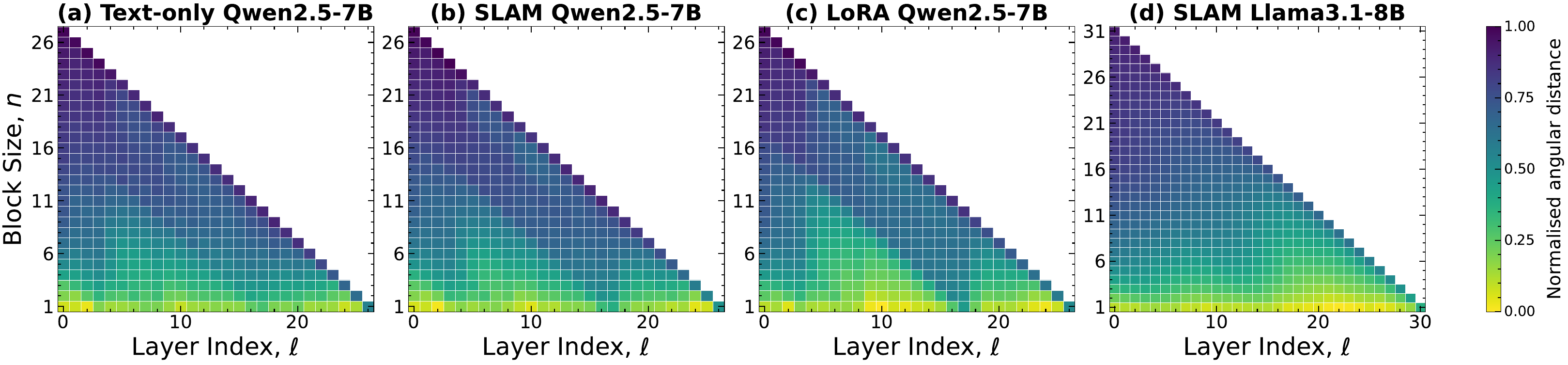}
  \caption{Angular distance between decoder layers $\ell$ and $\ell+n$, averaged over LibriSpeech dev-clean and dev-other. (a) Text-only input; (b) SLAM with frozen decoder; (c) SLAM with LoRA-adapted decoder; (d) SLAM Llama3.1-8B. 
  %Panels (a--c) share nearly identical structure, confirming that redundancy is inherited from the pretrained LLM; LoRA amplifies rather than disrupts the pattern. 
  %Deeper layers exhibit the highest similarity while the final layer remains maximally distant in all configurations.
  }
  \label{fig:heatmap_grid}
\end{figure*}

\begin{table}[!ht]
  \centering
  \caption{ASR WER (\%) before and after pruning using $\Delta_{\mathrm{WER}}$ threshold. The models correspond to the decoder+projector healing setup. \emph{drop} denotes the fraction of decoder layers removed. Loq. refers to the out-of-domain Loquacious dataset.}
  \label{tab:asr_wer_main}
  \begin{tabular}{lccccc}
    \hline
    & \% & \multicolumn{2}{c}{LibriSpeech} & Loq. \\
    \cline{3-4}\cline{5-5}
    Model & drop & clean & other & dev \\
    \hline
    Qwen2.5-7B      & --    & 2.01 & 4.77 & 12.21 \\
    \quad + Pruning & 28.6  & 2.36 & 4.83 & 14.91 \\
    \hline
    Qwen2.5-7B + LoRA     & --    & 1.70 & 3.56 & 10.99 \\
    \quad + Pruning     & 17.9  & 2.02 & 4.26 & 13.01 \\
                     \hline
    Qwen2.5-3B      & --    & 2.22 & 4.83 & 17.24 \\
    \quad + Pruning & 30.6  & 2.53 & 5.12 & 16.69 \\
                     \hline
    Qwen2.5-1.5B    & --    & 2.59 & 5.05 & 16.94 \\
    \quad + Pruning & 25.0  & 2.83 & 5.97 & 19.79 \\
    \hline
    Llama-3.1-8B    & --    & 2.65 & 5.03 & 20.50 \\
    \quad + Pruning  & 43.8  & 3.28 & 6.28 & 18.92 \\
                     \hline
    Llama-3.2-3B    & --    & 2.25 & 5.30 & 25.47 \\
    \quad + Pruning  & 39.3  & 2.71 & 6.32 & 25.72 \\
                     \hline
    Llama-3.2-1B    & --    & 2.61 & 5.52 & 28.05 \\
    \quad + Pruning  & 6.3   & 2.73 & 6.28 & 32.05 \\
    \hline
  \end{tabular}
\end{table}

Our understanding analysis proceeds in three stages: (i) tracing the origin of decoder redundancy by comparing angular distance heatmaps under different setups; (ii) studying the effect of the healing dynamics, and quantifying the redundancy on ASR; and (iii) exploring if the same overparameterisation extends to speech translation. Table~\ref{tab:asr_wer_main} reports baseline ASR models and pruned systems under $\Delta_{\mathrm{WER}}$ threshold.

\subsection{Origin of decoder redundancy}
\label{subsec:origin}

% The heatmaps in Figure~\ref{fig:heatmap_grid} show the angular distance between decoder layers for Qwen2.5-7B under three conditions: text-only input, SLAM with frozen decoder, and SLAM with LoRA-adapted decoder, alongside SLAM Llama3.1-8B for cross-architecture comparison.

\noindent \textbf{Understanding redundancy. }Panels (a) and (b) of Figure~\ref{fig:heatmap_grid} are nearly identical: the same blocks of layers exhibit low angular distance (average difference $\approx 0.007$), and the optimal pruning paths closely coincide. These results are consistent with redundancy being largely inherited from the pretrained LLM: speech routing via the projector does not substantially alter the measured layer-similarity structure. A practical consequence is that prunable layers can be identified from cheap text-only forward passes without training a SpeechLLM. Panel (d) shows that the same pattern is exhibited by Llama3.1-8B, confirming that the finding is generalised across architectures. Panel (c) shows that LoRA adaptation amplifies rather than disrupts this structure, with the average angular distance of the pruned blocks decreased from $0.36$ and $0.35$ for text and SLAM to $0.32$, indicating greater similarity in layer dynamics, though pruning robustness is not improved (Section~\ref{sec:pruning}). Across the analysed models, the smallest distances appear in deeper blocks, indicating that deeper layers are the most removable, while blocks including the final layer exhibit maximal distances, aligning with \cite{gromov2025unreasonableineffectivenessdeeperlayers}. Larger models also exhibit wider low-distance regions, suggesting greater excess capacity.

\subsection{Pruning redundant layers on speech recognition}
\label{sec:pruning}
%todo: discuss 3B as well + sometimes improvement + discuss loquacious 
\begin{figure*}[!t]
  \centering
  \includegraphics[width=\textwidth]{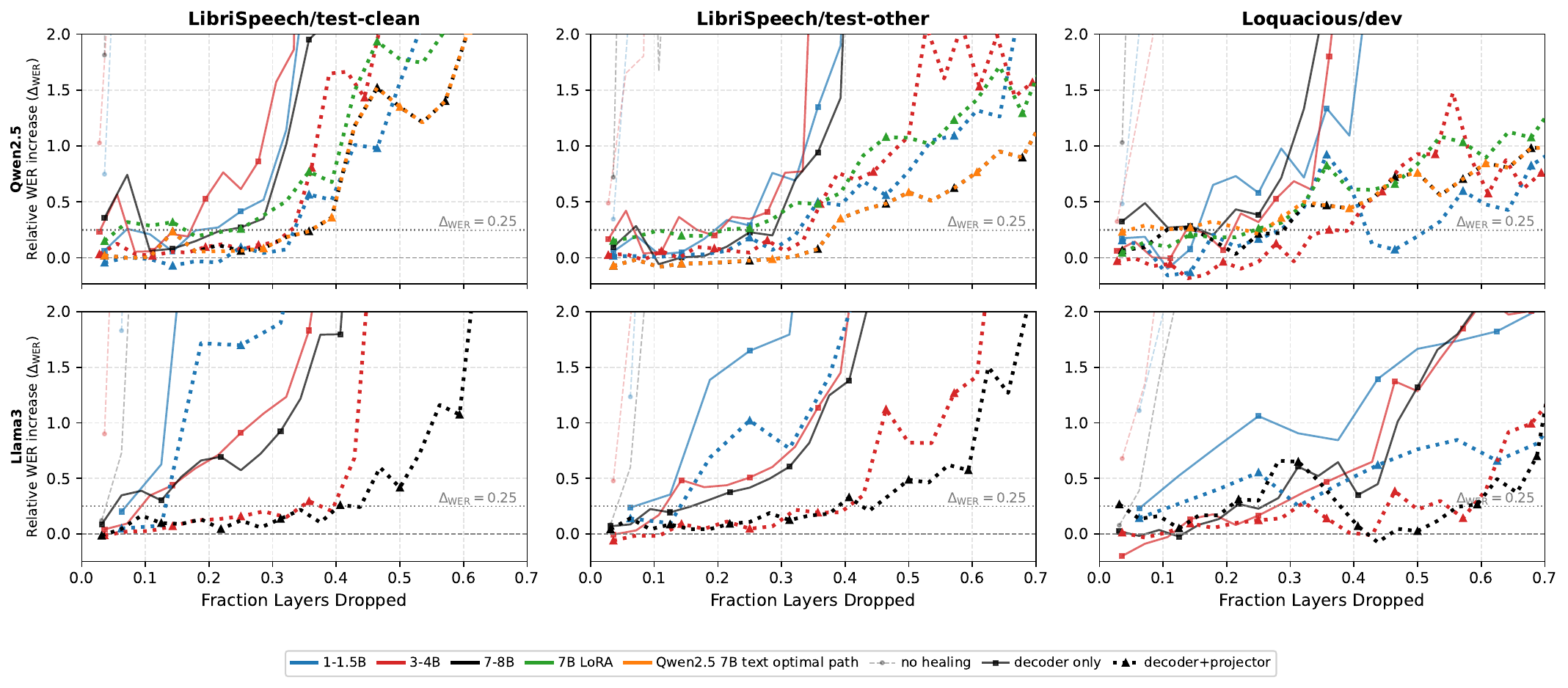}
  \caption{Relative WER degradation as a function of the fraction of decoder layers removed, for each evaluation set (columns) and model family (rows). The y-axis shows relative WER ($\Delta_{\mathrm{WER}}$) with respect to the unpruned baselines (Table~\ref{tab:asr_wer_main}); a value of $2.0$ means twice the baseline WER. The grey dashed line at $0.25$ marks the maximum allowed relative degradation threshold used in our analysis.}
  \label{fig:pruning_wer}
\end{figure*}

Having established that decoder redundancy is inherited from text and preserved under speech, we first analyse healing dynamics to determine a robust pruning recipe, and then quantify how much decoder capacity is actually needed for ASR.

\noindent \textbf{Healing dynamics.} To understand the respective roles of the decoder and projector in ASR performance, layers are progressively removed along the \emph{optimal pruning path} $\ell^*(n)$, and three healing strategies are compared: decoder-only, projector-only, and decoder+projector. Figure~\ref{fig:pruning_wer} shows the relative WER degradation as a function of pruning across our six models. Without healing, WER degrades sharply, often exceeding 50\% relative degradation. Healing only the decoder's receiving layer stabilises WER but still yields substantial degradation, whereas joint decoder+projector healing provides the best robustness. For Qwen2.5-7B on test-clean, pruning 28.6\% of layers yields 2.36\% WER with joint healing versus 5.93\% with decoder-only healing. Healing only the projector performs comparably to no healing, confirming that its limited capacity cannot compensate for removed layers. The gain from joint healing likely reflects projector realignment: pruning changes decoder dynamics, so the projector must be re-adapted. Following \cite{gromov2025unreasonableineffectivenessdeeperlayers}, the pruning mismatch is localised at the receiving layer $(\ell^*{+}n)$. Whereas they heal it by fine-tuning the MLP modules across the whole model, we restrict healing to that receiving block, which more directly isolates redundancy from added capacity.

\noindent \textbf{How many layers are needed for ASR?} Using the $\Delta_{\mathrm{WER}}$ criterion from Section~\ref{sec:experiments}, Table~\ref{tab:asr_wer_main} shows that 7--8B models maintain ASR performance within the acceptable relative degradation with only 63.8\% of their decoder layers, 3--4B models with 65.05\%, and 1--1.5B models with 86.5\%. Larger decoders carry proportionally more excess capacity. While layers can be pruned beyond our threshold as shown in Figure~\ref{fig:pruning_wer}, performance quickly falls below that of the unpruned smaller counterparts. Importantly, the relative degradation depends strongly on the initial performance: a strong ASR system (e.g.\ Qwen2.5-7B) may exhibit stronger degradation while still achieving better absolute WER than another model (e.g.\ Llama3.1-8B). Interestingly, in Figure~\ref{fig:pruning_wer}, pruning occasionally improves out-of-domain WER. This is attributed to a regularisation effect: the projector is originally aligned on LibriSpeech, and since pruning alters the decoder dynamics, healing forces a new alignment that may be less domain-specific. Despite amplifying representation similarity, LoRA reduces pruning tolerance (17.9\% vs 28.6\%), suggesting that adaptation introduces functional dependencies not captured by angular distance. Finally, as a practical consequence, removing 40\% of Llama3.1-8B decoder layers yields a 35\% wall-clock speedup and reduces peak GPU memory from 15.72 to 10.37~GiB (H100 80~GiB).

\noindent \textbf{Cross-modal path transfer.} We compare text-only and speech-derived optimal pruning paths on Qwen2.5-7B. The two paths yield near-identical WER at every pruning level: in Figure~\ref{fig:pruning_wer}, the orange curve closely tracks the speech-derived path across all evaluation sets. This indicates that layers dispensable for speech are also dispensable for text, and that prunable layers can be identified from text-only forward passes.

\subsection{Generalisation to speech translation}

\begin{table}[t]
  \centering
  \caption{AST BLEU scores on CoVoST2 Fr$\rightarrow$En and En$\rightarrow$De for Qwen2.5-7B using the AST-optimal and ASR-optimal pruning paths. Higher is better.}
  \label{tab:ast_bleu}
  \begin{tabular}{rcccc}
    \hline
    & \multicolumn{2}{c}{AST opt.\ path} & \multicolumn{2}{c}{ASR opt.\ path} \\
    \cline{2-3}\cline{4-5}
    \% drop & Fr$\rightarrow$En & En$\rightarrow$De & Fr$\rightarrow$En & En$\rightarrow$De \\
    \hline
    -- & 39.03 & 27.66 & 39.03 & 27.66 \\
    10.7\% & 37.23 & 27.46 & 37.23 & 27.40 \\
    21.4\% & 36.09 & 25.57 & 36.09 & 25.55 \\
    32.1\% & 36.12 & 25.40 & 36.12 & 25.40 \\
    42.9\% & 32.87 & 17.08 & 32.93 & 17.07 \\
    \hline
  \end{tabular}
\end{table}

We next assess whether the ASR findings transfer to AST across two non-English directions (Fr$\rightarrow$En, En$\rightarrow$De) and a different speech encoder (Whisper Large v3), testing transferability beyond the English ASR setting. If redundancy is primarily inherited from the pretrained LLM backbone (Section~\ref{subsec:origin}), both pruning tolerance and optimal paths should remain consistent across tasks, languages, and encoders.

\noindent \textbf{How many layers are needed for AST?} As in ASR, pruning based on $\ell^*(n)$ without healing causes sharp BLEU degradation, while healing both decoder and projector yields the strongest robustness. Table~\ref{tab:ast_bleu} reports absolute BLEU as a function of the fraction of layers dropped. Using the $\Delta_{\mathrm{BLEU}}$ threshold from Section~\ref{sec:experiments}, 32.1\% of layers are removable on Fr$\rightarrow$En and En$\rightarrow$De while preserving speech translation capabilities. This robustness is comparable to the ASR case in Table~\ref{tab:asr_wer_main}.

\noindent \textbf{Cross-task path transfer.} Section~\ref{sec:pruning} shows that the text-only and speech-derived pruning paths coincide for ASR. We now evaluate whether this path-level agreement also holds for AST, which differs in task objective, target language, and speech encoder. The AST-optimal path closely matches the ASR-optimal path: Table~\ref{tab:ast_bleu} shows that applying the ASR-derived path to AST yields virtually identical BLEU at every pruning level. This indicates that the same decoder layers are redundant across the tested ASR and AST settings. A single pruned backbone can therefore support multiple tasks with task-specific adapters in this setup. Since the redundancy structure appears modality- and task-agnostic, these findings may extend to other tasks where deeper pretrained LLM layers are redundant, like question answering and reasoning tasks as suggested by  \cite{gromov2025unreasonableineffectivenessdeeperlayers}.

\section{Conclusions}

We investigated the problem of excess capacity in SpeechLLMs across six models. Using redundancy as a proxy, we first showed that this phenomenon is largely inherited from the pretrained LLM backbone, with similar prunable layer blocks across text and speech. We then analysed post-pruning healing dynamics and found that jointly adapting the projector and decoder is critical for pruning robustness. Under this recipe, 7--8B models retain good ASR performance with only $\sim$60\% of decoder layers, while other model scales also exhibit excess redundancy. We further showed that these findings transfer to AST: comparable fractions of layers are removable, and ASR- and AST-redundant prunable blocks along the optimal pruning paths closely coincide. Overall, our results suggest that decoder redundancy is a broader phenomenon that appears to be modality- and task-agnostic, allowing the possibility to deploy a single pruned multi-task SpeechLLM decoder. Extending to more model families, datasets, languages and tasks is left to future work.

\section{Generative AI Use Disclosure}
Large Language Models (e.g. ChatGPT, Claude) were only used to ensure linguistic clarity in specific sections.

\bibliographystyle{IEEEtran}
\bibliography{mybib}
\end{document}